\documentclass{article}

\usepackage[final]{neurips_2022}

\usepackage[utf8]{inputenc} 
\usepackage[T1]{fontenc}    
\usepackage{hyperref}       
\usepackage{url}            
\usepackage{booktabs}       
\usepackage{amsfonts}       
\usepackage{nicefrac}       
\usepackage{microtype}      
\usepackage{xcolor}         
\usepackage[pdftex]{graphicx}
\usepackage{multirow}
\usepackage{enumitem}

\usepackage{float}
\restylefloat{table}
\usepackage[bottom]{footmisc}
\newcommand{\mOne}{Meg1.3b}
\newcommand{\mEight}{Meg8.3b}
\newcommand{\mTwenty}{Meg22b}

\usepackage{tcolorbox}
\definecolor{lightBlue}{rgb}{0.8, 0.9, 1.0}
\definecolor{lightRed}{rgb}{1.0, 0.90, 0.90}

\newtcbox{\bluebox}{on line, box align=base, colback=lightBlue,colframe=white,size=fbox,arc=3pt, before upper=\strut, top=-2pt, bottom=-4pt, left=-2pt, right=-2pt, boxrule=0pt}

\newtcbox{\redbox}{on line, box align=base, colback=lightRed,colframe=white,size=fbox,arc=3pt, before upper=\strut, top=-2pt, bottom=-4pt, left=-2pt, right=-2pt, boxrule=0pt}

\newcommand{\dashifted}{\raisebox{0.5\depth}{\tiny$\downarrow$}}
\newcommand{\upshifted}{\raisebox{0.5\depth}{\tiny$\uparrow$}}

\newcommand{\dar}[1]{{\scriptsize\redbox{\dashifted{#1}}}}

\newcommand{\uab}[1]{{\scriptsize\bluebox{\upshifted{#1}}}}

\newcommand{\caltech}{$^1$}
\newcommand{\nvidia}{$^2$}

\title{Can You Label Less by Using Out-of-Domain Data? \\Active \& Transfer Learning with Few-shot Instructions}

\author{Rafal Kocielnik$^*$\caltech{}, Sara Kangaslahti$^*$\caltech{}$^,$\nvidia{}, Shrimai Prabhumoye\nvidia{}, Meena Hari\caltech{}, \\ 
\bf R. Michael Alvarez\caltech{}, 
Anima Anandkumar\caltech{}$^,$\nvidia{}\\
  \caltech{}California Institute of Technology, \nvidia{}NVIDIA \\
  \texttt{\{rafalko@caltech.edu, skangasl@caltech.edu\}} \\}

\begin{document}

\maketitle
\def\thefootnote{*}\footnotetext{Both authors contributed equally to this research}

\begin{abstract}
  Labeling social-media data for custom dimensions of toxicity and social bias is challenging and labor-intensive. Existing transfer and active learning approaches meant to reduce annotation effort require fine-tuning, which suffers from over-fitting to noise and can cause domain shift with small sample sizes. In this work, we propose a novel Active Transfer Few-shot Instructions (ATF) approach which requires no fine-tuning. ATF leverages the internal linguistic knowledge of pre-trained language models (PLMs) to facilitate the transfer of information from existing pre-labeled datasets (\emph{source-domain task}) with minimum labeling effort on unlabeled target data (\emph{target-domain task}). Our strategy can yield positive transfer achieving a mean AUC gain of 10.5\% compared to no transfer with a large 22b parameter PLM. We further show that annotation of just a few \emph{target-domain samples} via active learning can be beneficial for transfer, but the impact diminishes with more annotation effort (26\% drop in gain between 100 and 2000 annotated examples). Finally, we find that not all transfer scenarios yield a positive gain, which seems related to the PLMs initial performance on the \emph{target-domain task}.
\end{abstract}

\begin{figure}[t]
  \centering
  \includegraphics[width=\linewidth]{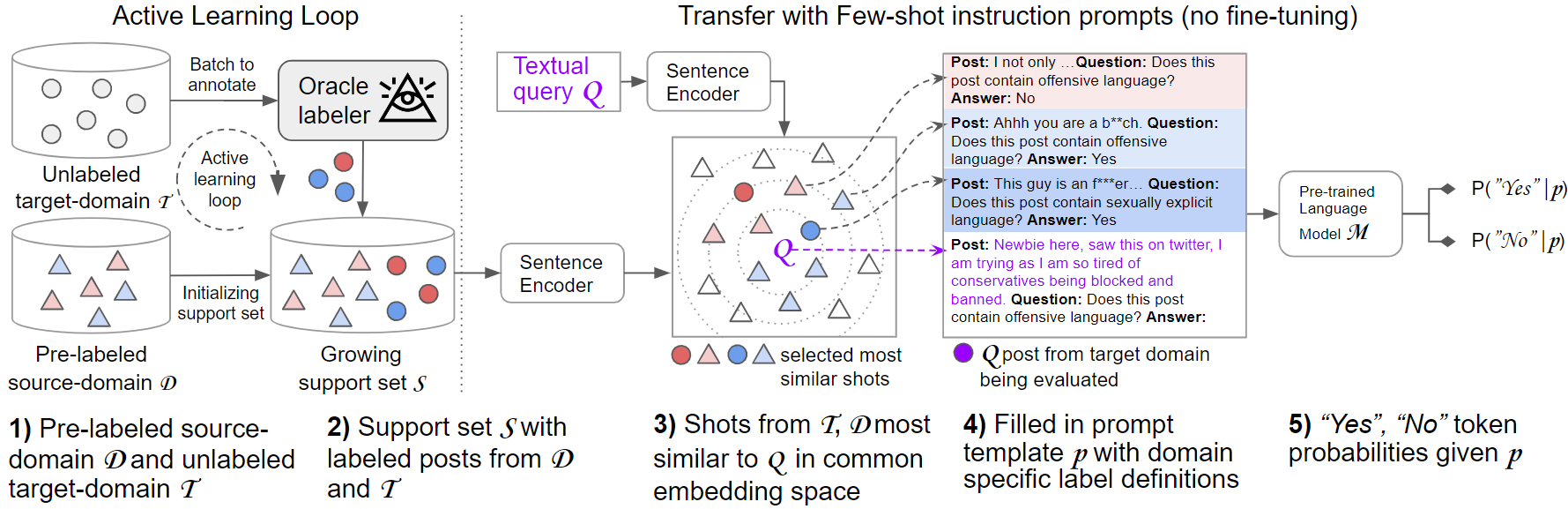}
  \vspace{-6.0pt}
  \caption{Overview of our approach: We populate a support set $\mathbf{S}$ with existing dataset $\mathbf{D}$ labeled under the \emph{source-domain} definition. In an active learning loop, we further add to $\mathbf{S}$ a small number of posts labeled by an oracle (annotation effort) under the \emph{target-domain} definition from a large unlabeled \emph{target-domain} dataset $\mathbf{T}$. We use a sentence encoder to project the textual posts from $\mathbf{S}$ and unlabeled query $\mathbf{Q}$ to the same embedding space. We use a cosine similarity metric to select the class-balanced shots most similar to $\mathbf{Q}$. The text of the selected shots, their labels, and the labeling definitions (corresponding to the original labeling dimension) are used to fill in an instruction template $\mathbf{p}$, and finally passed to the pre-trained LM to make a prediction based on the conditional token probability for each class. The novelty of our method stems from: 1) populating the support set $\mathbf{S}$ with a mixture of source and target domain posts, and 2) communicating the original labeling definitions as part of an instruction template $\mathbf{p}$ to allow the model to relate the two.}
  \label{fig:architecture_overview}
\end{figure}

\section{Introduction}
\label{sec:introduction}

While real-world social media data is abundant, its annotation for important issues such as toxicity, hate-speech, and various facets of social bias is inherently challenging \cite{patton2019annotating, modha2020tracking}. These challenges stem from the sheer number of nuanced and evolving dimensions \cite{liu2019finding} as well as inherent ambiguities in interpretation \cite{chen2018using} leading to noisy labeling \cite{wang2016sentiment}. Active and transfer learning approaches offer an ability to lower annotation effort by intelligently selecting the most informative examples to annotate \cite{farinneya2021active} or by using existing labeled datasets \cite{zhuang2020comprehensive}. However, most active learning approaches usually yield too few samples (on the order of hundreds) to feasibly fine-tune large deep-language models \cite{wang2016cost, kasai2019low}. In terms of transfer, fine-tuning on out-of-domain data can lead to detrimental domain shift \cite{ma2019domain}. Furthermore, fine-tuning can also lead to over-fitting, especially in the case of smaller train sets, and to catastrophic forgetting of knowledge present in the pre-trained model \cite{fatemi2021improving}. Hence, prior work mostly did not use PLMs in this setup\cite{farinneya2021active, zhao2021active}.

\textbf{Our Approach:} In this work, we propose the use of few-shot instructions (textual prompts) with PLMs as a fine-tuning-free alternative. Our approach can be effective with few samples \& is robust against social-media inherent labeling noise \cite{prabhumoye2021few}, which is not well handled by fine-tuning-based approaches \cite{song2022learning}. We propose an Active Transfer Few-shot Instructions (ATF) method for combining active learning (for selecting fewer samples to label) with transfer learning (for leveraging existing labeled datasets) under few-shot instructions with PLMs. Our method leverages the capacity of PLMs to 1) learn from a few examples in a few-shot fashion without fine-tuning and 2) transfer task knowledge from datasets already labeled under different definitions to further reduce the need for costly annotation. We experiment with transfer scenarios on 3 datasets across 8 labeling dimensions provided by crowd-sourcing and an existing state-of-the-art commercial tool - Perspective API \cite{PerspectiveAPI:online}.

\textbf{Prior work:} Several recent works studied the use of few-shot instructions and in-context learning for lowering annotation efforts. They, however, focused either on sample-selection strategies  \cite{su2022selective, yu2022cold} or improving few-shot performance of smaller models \cite{wang2021entailment, gao2020making, mishra2021reframing}, but did not study transfer from existing pre-labeled datasets. Several works also employed various fine-tuning approaches in low resource settings \cite{kasai2019low, lee2021good}. Works attempting transfer with PLMs again turn to fine-tuning in a text-to-text format \cite{raffel2020exploring} or attempt transfer in a few-shot setting by framing the problem as \textit{instruction tuning}, where PLMs are fine-tuned on a collection of datasets described via instructions \cite{wei2021finetuned, min2021metaicl}. Our work is different from all these approaches, as we focus on transfer from prelabeled external datasets via few-shot instructions without fine-tuning under a basic random active learning setting. In fact, our ATF method can be used in synergy with better sample selection strategies proposed in \cite{su2022selective, yu2022cold}.

\begin{figure}[t]
  \centering
  \begin{minipage}{0.58\linewidth}
      \includegraphics[width=\linewidth]{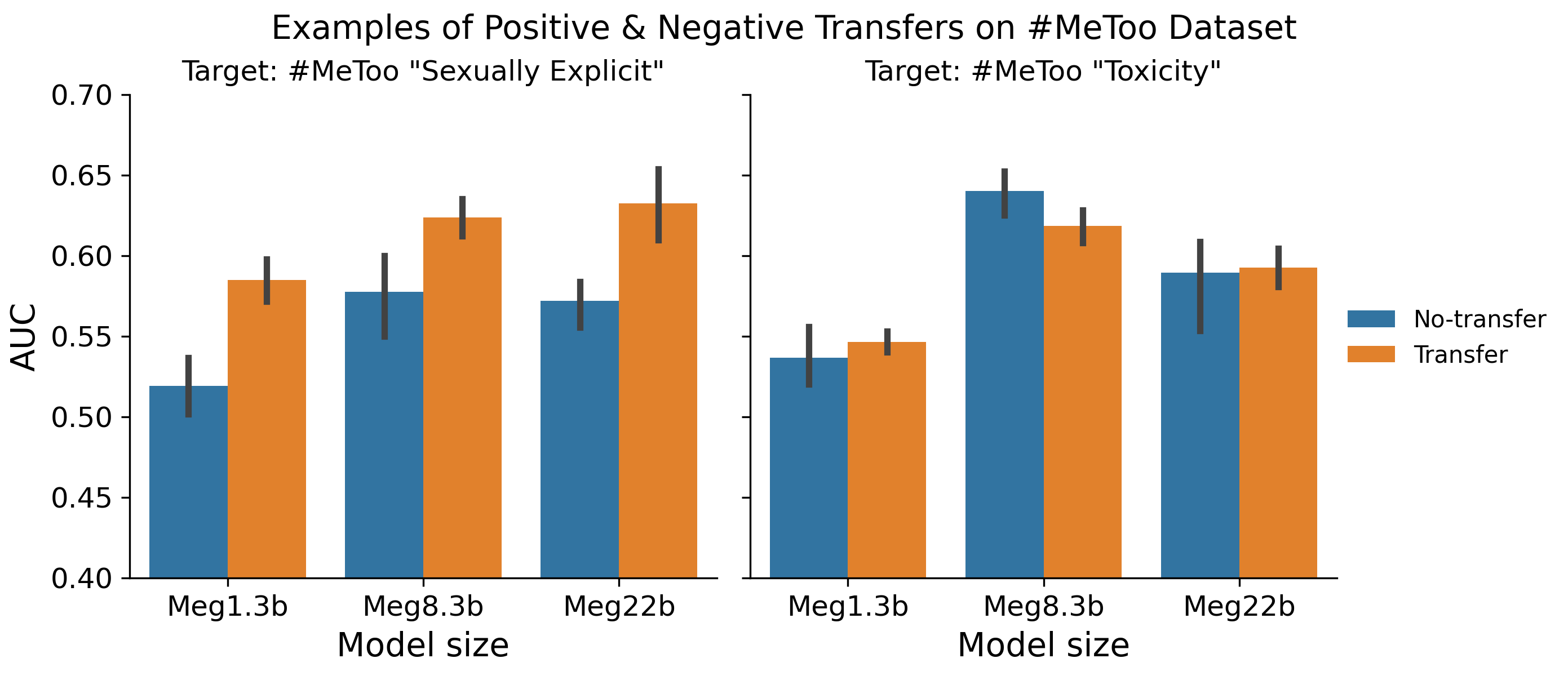}
      \vspace{-6.0pt}
      \caption{Example of positive and negative transfer from existing datasets (SBIC \& HASOC) with 6 pre-labeled source dimensions (Table \ref{fig:main_transfer}) to Perspective API labeled dimensions (``Sexually Explicit'' and ``Toxicity'') on a real-world dataset (\#MeToo). Positive transfer is retained across model sizes, while negative transfer starts as a minor gain that turns into a loss as model size increases.}
      \label{fig:main_transfer}
  \end{minipage}
  \hfill
  \begin{minipage}{0.385\linewidth}
      \includegraphics[width=\linewidth]{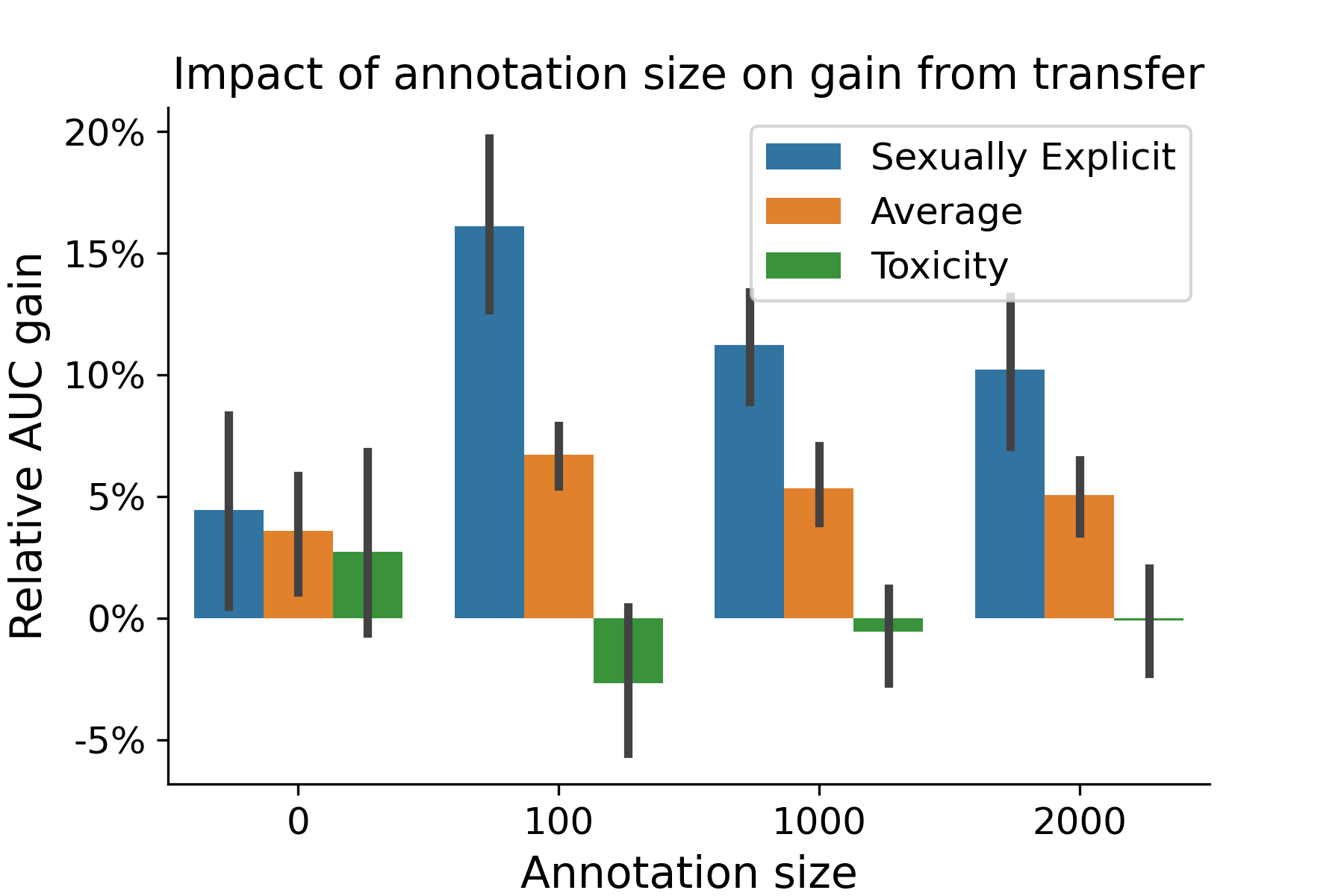}
      \vspace{-6.0pt}
      \caption{Relative averaged impact of target annotation size on transfer effectiveness. We can see that annotating small number of target domain samples via active learning can benefit transfer, but the relative gain diminishes as more target samples are annotated.}
      \label{fig:sample_size_transfer}
  \end{minipage}
  
\end{figure}

\textbf{Findings:} We find that using pre-labeled \emph{source-domain} data can help improve classification results when very few  examples from the \emph{target-domain} are labeled. We further observe two scenarios: positive and negative transfer (\ref{fig:main_transfer}). When the positive transfer occurs, it leads to high AUC gains that are consistently sustained across model sizes (12.94\% for 1.3b to 10.49\% for 22b) and annotation sizes (16.02\% for 100 to 10.19\% for 2000 annotations). Negative transfer leads to small inconsistent gains that can turn into losses with larger model size (1.91\% for 1.3b to -3.30\% for 8.3b). We observe that as few as 100 target domain annotations can aid transfer increasing mean gain from 3.64\% to 6.73\%. However, the transfer gain diminishes with more labeled examples from the \emph{target-domain} (at the expense of annotation effort), falling from 6.73\% to 4.97\% (Figure \ref{fig:sample_size_transfer}). Finally, we investigate the reasons behind positive and negative transfers and find that the higher the AUC the PLM can achieve with only \emph{target-domain} data, the less gain can be expected from the transfer ($\rho$\footnote{Pearson correlation coefficient}=$-0.66$). 

\textbf{Contributions:} In this work we offer the following contributions:
\begin{itemize}[leftmargin=*]
\item Novel adaptation of few-shot instructions to facilitate transfer learning without fine-tuning on PLMs under limited annotation resources - Active Transfer Few-shot Instructions (ATF).

\item Insights into the reasons for negative \& positive transfers when attempting transfer learning with few-shot instructions and no fine-tuning.
\end{itemize}

\section{Methodology}
\label{sec:methodology}

\paragraph{Few-shot Instructions:} We adapt the few-shot prompting approach detailed in \cite{prabhumoye2021few} as shown in Figure \ref{fig:architecture_overview}; that is, given a query post from the test set, we present PLM with an input in the form $\mathbf{p} = [``\mathtt{Post:}"; \mathbf{Q}; ``\mathtt{Question:}"; \mathbf{d}; ``\mathtt{Answer:}"]$, where we concatenate the tags \textit{Post}, \textit{Question} and \textit{Answer}. We calculate the probabilities of the tokens \textit{Yes} and \textit{No} in the following manner: $p_{\mathcal{M}}(\mathtt{``Yes"} | \mathbf{p})$ and $p_{\mathcal{M}}(\mathtt{``No"} | \mathbf{p})$. The token that has the higher probability is considered the prediction for the post. 
We define a \emph{support set} $\mathbf{S}$ as a set of labeled examples (which contains a mix of source and target domain posts) from which the shots for few-shot instructions are selected.
To select the 32 shots (same as in \cite{prabhumoye2021few}), we sample class balanced exemplars from $\mathbf{S}$ based on their semantic similarity with the query post $\mathbf{Q}$, computed using cosine similarity in the common embedding space encoded using Term Frequency-Inverse Document Frequency (TF-IDF) representation \cite{TfIdfSciKit:online}. We present the PLM with these exemplars as in-context samples using the same structure as $\mathbf{p}$. To adapt this approach to the transfer learning setting, we present the shots with the definition under which they were originally labeled (Table \ref{tab:labeling_definitions_table}).

\paragraph{Transfer learning setup:} For the baselines, the support set $\mathbf{S}$ consists of only a few target domain examples selected from a large unlabeled target domain dataset $\mathbf{T}$ and labeled by an oracle (Figure \ref{fig:architecture_overview}). However, for the transfer learning experiments, we augment these target domain examples with the entirety of the pre-labeled source domain training dataset $\mathbf{D}$. Thus, shots are selected from this augmented support set $\mathbf{S}$.
    
\paragraph{Active learning setup:} We utilize an active learning setup to evaluate the performance of the model with varying target domain annotation sizes (Figure \ref{fig:architecture_overview}). To do so, we use the unlabeled pool scenario, in which we have a small amount of labeled target domain data and a large unlabeled target domain dataset $\mathbf{T}$. We simulate this scenario by first randomly sampling $100$ examples from $\mathbf{T}$ to be labeled by an oracle. In subsequent iterations, we randomly sample from the remaining unlabeled data in $\mathbf{T}$ to provide the model with \textit{support set} $\mathbf{S}$ with $1000$ and $2000$ labeled target domain examples. As we are randomly sampling from a large dataset, we repeat the entire pipeline five times for each experiment to ensure that our results are stable. We fix the random seed for each iteration of the pipeline for reproducibility and consistency across experiments.

\paragraph{Models:} For each task, we use off-the-shelf PLMs. We utilize the Megatron 1.3B parameter model (\mOne), Megatron 8.3B parameter model (\mEight), and Megatron 22B parameter model (\mTwenty), all of which have been pre-trained using the toolkit in \cite{shoeybi2019megatron}. 

\paragraph{Metrics:} We evaluate the performance of the model using the area under the curve (AUC) \cite{AUCSciKit:online}, which measures how well a classifier can discriminate between classes, as the tasks we consider are all binary but have varying percentages of positive ("Yes") labels.

\paragraph{Perspective API labeling:} To perform transfer experiments in a controlled manner, we obtained an external and consistent set of labels related to hate speech and toxicity for all our data. We used Perspective API, a state-of-the-art pretrained toolkit \cite{PerspectiveAPI:online}, which has been used in \cite{schick2021self, hartvigsen2022toxigen}. While Perspective API has been reported to have limitations in relation to biased classification and limited labeling dimensions \cite{bender2021dangers}, it still represents a good off-the-shelf baseline.

\begin{table*}[t]
\centering
\small{
\begin{tabular}{@{}l l @{\hskip 0.10in} r r r  r r r @{}}
\textbf{Source} & \textbf{Target} & \multicolumn{3}{c}{\textbf{Megatron1.3b}} & \multicolumn{3}{c}{\textbf{Megatron22b}}\\
\toprule
\multicolumn{2}{c}{Annotation size} & AUC@100 & @1k & @2k & AUC@100 & @1k & @2k\\
\midrule
None & \multirow{2}{0.5in}{MeToo ``Sexually Explicit''} & 54.0 & 49.5 & 53.4 & 57.7 & 57.6 & 58.7\\
SBIC "Lewd" &  & \uab{17\%}58.2 & \uab{5\%}55.9 & \uab{18\%}59.8 & \uab{7\%}61.9 & \uab{9\%}62.9 & \uab{10\%}64.7\\
SBIC "Group" &  & \uab{25\%}61.7 & \uab{12\%}59.8 & \uab{6\%}53.8 & \uab{17\%}67.6 & \uab{10\%}63.0 & \uab{8\%}63.6\\
SBIC "Intent" &  & \uab{24\%}61.3 & \uab{13\%}60.3 & \uab{17\%}59.5 & \uab{6\%}61.3 & \uab{10\%}63.6 & \uab{15\%}67.3\\
SBIC "Offensive" & & \uab{26\%}62.5 & \uab{16\%}61.8 & \uab{14\%}57.8 & \uab{20\%}69.0 & \uab{20\%}69.1 & \uab{14\%}67.1\\
HASOC "HOF" & & \uab{23\%}61.0 & \uab{5\%}56.6 & \uab{18\%}60.1 & \uab{22\%}70.3 & \uab{18\%}67.8 & \uab{12\%}66.0\\
HASOC "Target" & & \uab{28\%}63.2 & \uab{17\%}62.3 & \uab{22\%}61.9 & \uab{12\%}64.5 & \uab{17\%}67.6 & \uab{9\%}64.0\\
\midrule
None & \multirow{2}{0.5in}{MeToo ``Toxicity''} & 51.5 & 53.5 & 53.3 & 61.0 & 60.8 & 60.5\\
SBIC "Lewd" &  & \uab{11\%}57.1 & \uab{0\%}53.6 & \uab{4\%}55.4 & \dar{5\%}57.7 & \dar{1\%}59.9 & \dar{3\%}58.7\\
SBIC "Group" &  & \dar{1\%}51.2 & \uab{1\%}54.1 & \dar{2\%}52.3 & \dar{7\%}56.5 & \dar{5\%}57.9 & \dar{5\%}57.4\\
SBIC "Intent" & & \uab{6\%}54.8 & \uab{2\%}54.4 & \uab{0\%}53.6 & \dar{6\%}57.5 & \dar{12\%}53.6 & \dar{8\%}54.8\\
SBIC "Offensive" &  & \uab{7\%}55.3 & \dar{1\%}53.1 & \dar{0\%}53.2  &  \dar{3\%}59.2 & \dar{5\%}57.8 & \dar{3\%}58.4\\
HASOC "HOF" & & \uab{8\%}55.7 & \dar{0\%}53.2 & \uab{7\%}56.9 & \dar{6\%}57.4 & \uab{2\%}62.3 & \uab{5\%}63.4\\
HASOC "Target" & & \uab{4\%}53.8 & \uab{2\%}54.5 & \uab{6\%}56.3 & \dar{6\%}57.3 & \uab{3\%}62.4 & \uab{8\%}65.2\\
\bottomrule

\end{tabular}
}
\vspace{1.0em}
\caption{Absolute AUC and relative gain or loss compared to no transfer for 2 transfer scenarios with targets of \#MeToo dataset dimensions ``Sexually Explicit'' and ``Toxicity'' as labeled by Perspective API. The sources are all pre-labeled dimensions provided with SBIC \& HASOC datasets. Target ``Sexually Explicit'' represents a positive transfer, where gains are sustained across the annotation set and model sizes. Target ``Toxicity'' represents a negative transfer scenario, where gains are small and inconsistent in the smaller model and quickly turn into losses with a larger model.}
\vspace{-1.0em}
\label{tab:transfer_results_table}
\end{table*}

\section{Datasets and Results}
    
\subsection{Datasets}
\label{sec:datasets}

We use three datasets and a total of eight labeling dimensions for our experiments: SBIC~\cite{sap-etal-2020-social}, HASOC~\cite{mandl-etal-2019-hasoc} and \#MeToo~\cite{srikanth2021dynamic}. We report correlations between labeling dimensions for these datasets in Figure \ref{fig:datasets_corrs}, Appendix \ref{sec:correlations_appendix} and an estimate of the distributional difference between them in Figure \ref{fig:datasets_distribution_difference}, Appendix \ref{sec:separability_similarity_appendix}.

\paragraph{Social Bias Frames (SBIC):}
\label{sec:ds-sbic}
This dataset~\cite{sap-etal-2020-social} contains 34k documents in the training set labeled under categories in which people project social biases and stereotypes onto others. We use four binary classification tasks, which were all labeled by crowd-workers. These tasks have the following labels and definitions: (1) offensive ($57.5\%$ positive labels): whether a post could be considered "offensive" to anyone, (2) intent ($53.1\%$ positive): whether the perceived motivation of the author was to offend, (3) lewd ($9.6\%$ positive): whether a post contains sexual references, (4) group ($41.1\%$ positive): whether a post is offensive toward a group. We also use Perspective API to label: (1) toxicity ($39.1 \%$ positive): whether a post is rude, disrespectful, or unreasonable and likely to make people leave a discussion and (2) sexually explicit ($22.3\%$ positive): whether a post contains references to sexual acts, body parts, or other lewd content.

\paragraph{Hate Speech and Offensive Content Identification (HASOC):}
\label{sec:ds-hasoc}
The HASOC dataset~\cite{mandl-etal-2019-hasoc} contains documents from Twitter and Facebook, which were developed for identifying hate speech and offensive content. The dataset contains documents in three languages, but we use only the English tasks, which consist of 6k documents. We utilize the two binary classification tasks in the English dataset, which is human-labeled. The tasks are defined as follows: (1) HOF ($25.0\%$ positive): whether a post contains hate, offensive, or profane content, (2) Target ($21.3\%$ positive) whether a post contains an insult (targeted or untargeted). We also label this dataset under the toxicity ($25.6\%$ positive) and sexually explicit ($16.9\%$ positive) Perspective API tasks.

\paragraph{\#MeToo Twitter Dataset:}
\label{sec:ds-metoo}
The \#MeToo women's rights movement became popular very quickly on Twitter after gaining exposure from a tweet by Alyssa Milano in October 2017. There were many anecdotal stories that women who participated in the \#MeToo movement on Twitter were subjected to harassment and trolling. Thus, in this paper we use Twitter data from the \#MeToo movement to investigate toxicity and harassment directed at movement participants. We utilize a dataset created by collecting the tweets from January $2017$ to September $2019$ that contain \#MeToo related keywords ~\cite{srikanth2021dynamic}. This \#MeToo dataset consists of $7.55$ million documents after preprocessing. We label this dataset using the toxicity ($18.8 \%$ positive) and sexually explicit ($18.6\%$ positive) dimensions from Perspective API.

\begin{figure}[t]
  \centering
  \begin{minipage}{0.59\linewidth}
      \includegraphics[width=\linewidth]{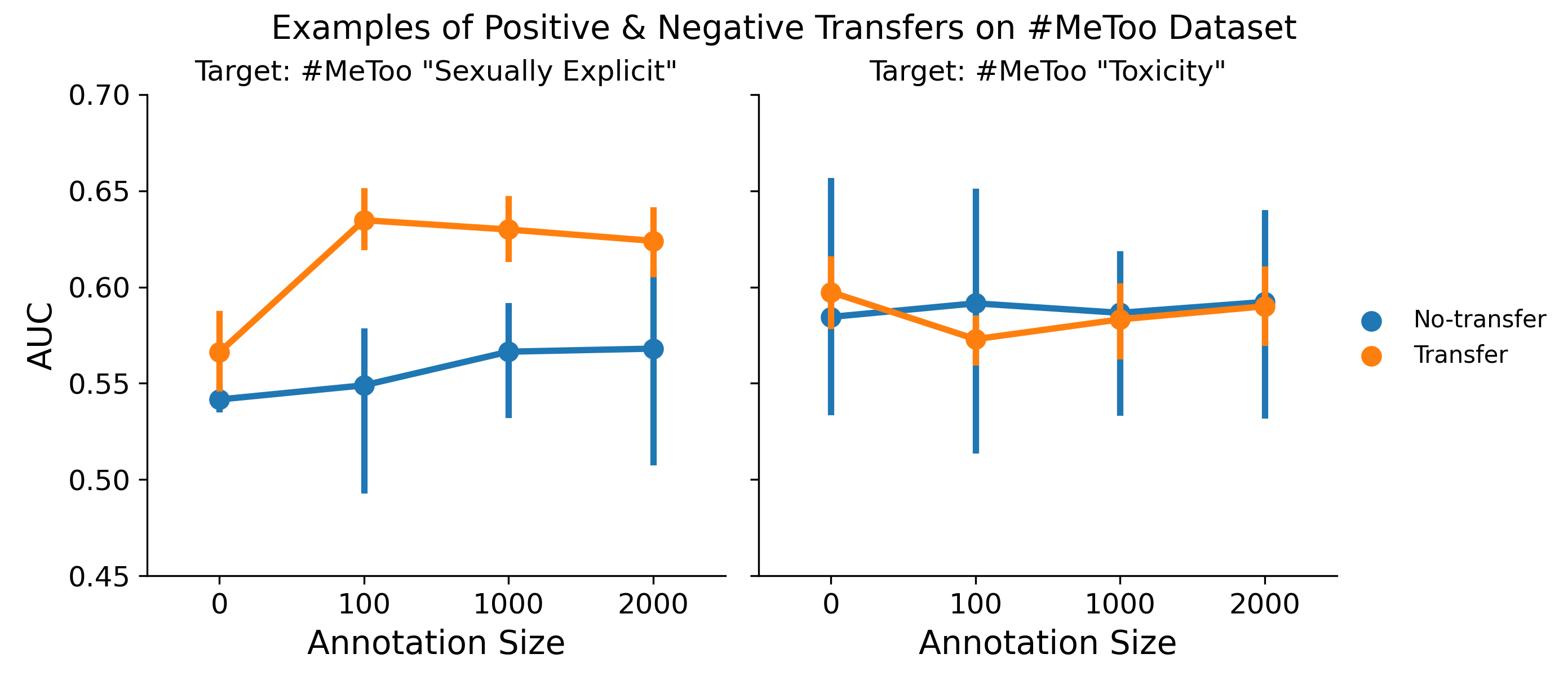}
      \vspace{-11.0pt}
      \caption{Comparison between transfer and no-transfer scenarios across the target domain annotation sizes split by positive (\#MeToo ``Sexually Explicit'') and negative (\#MeToo ``Toxicity'') transfer scenarios. We can see that in the positive transfer scenario, the gain from transfer occurs at ever annotation size, but seem optimal with as few as 100 annotated target samples after which id diminishes.  }
      \label{fig:by_annot_size}
  \end{minipage}
  \hfill
  \begin{minipage}{0.38\linewidth}
    \includegraphics[width=\linewidth]{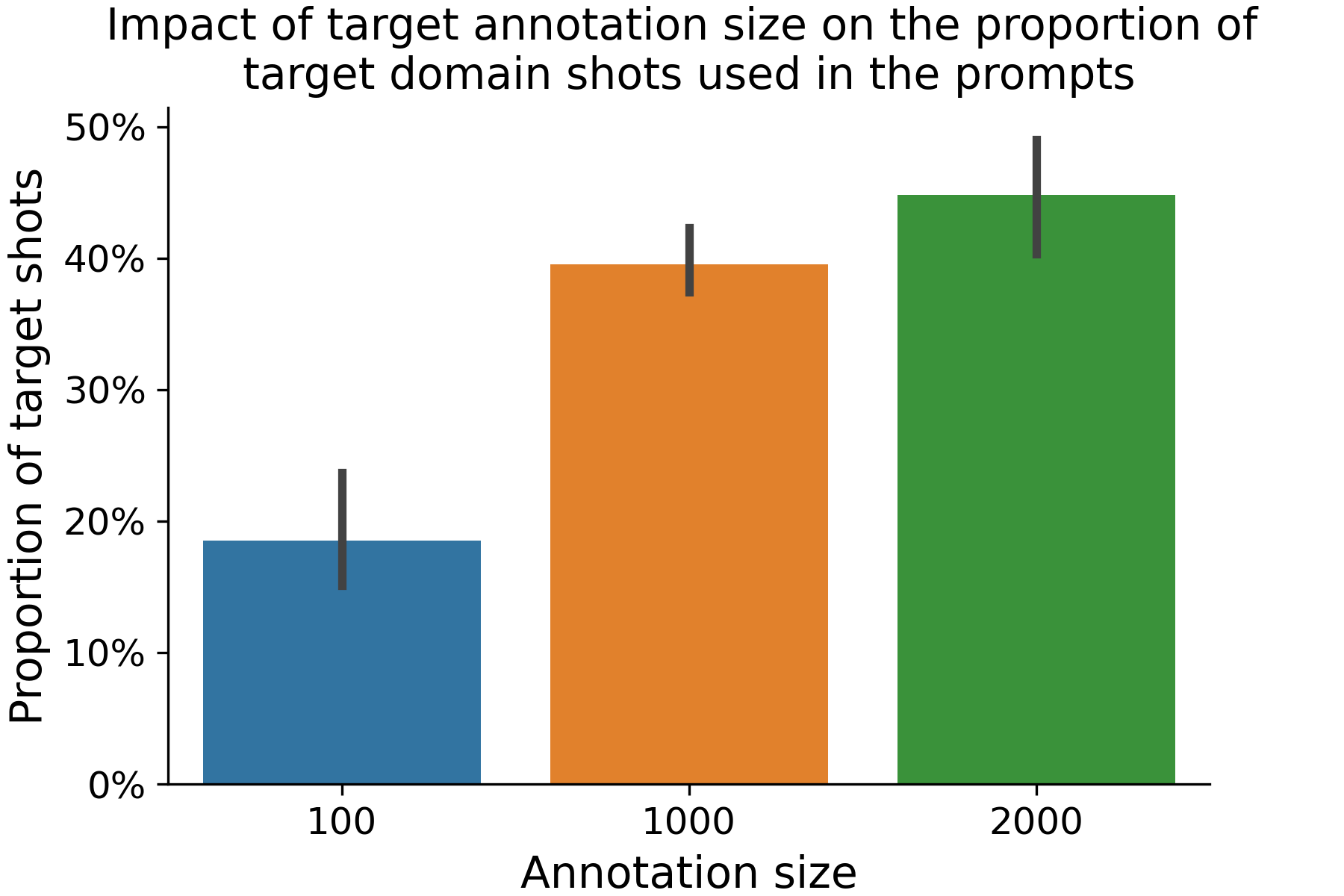}
    
      \caption{Impact of target domain annotation size on the proportion of target examples selected as shots for few-shot prompts. We can see that this proportion increases as target domain examples are more likely to be picked as shot by semantic similarity metric.}
      \label{fig:shot_ratio}
      
    \end{minipage}

\end{figure}

\subsection{Results}
\label{sec:results}

\paragraph{Transfer Effectiveness:} The results of transfer experiments with two model sizes \mOne{} and \mTwenty{} for two target dimensions from the \#MeToo dataset (``Sexually Explicit'' and ``Toxicity'') and 6 source dimensions from SBIC and HASOC are presented in Table \ref{tab:transfer_results_table}. All the results are presented as absolute AUC scores under growing target annotation size.  Next to each AUC score we show the relative gain or loss compared to the no-transfer baseline using only the annotated target samples. We also used an intermediate-sized model (\mEight) and target annotation size of 0, which are not shown in the table, but averaged results are presented in Figure \ref{fig:main_transfer} across model sizes and in Figure \ref{fig:by_annot_size} across annotation sizes. We include these results in our summaries.

The examples represent positive and negative transfer scenarios. In the positive transfer (\#MeToo ``Sexually Explicit''), it can be seen that the gains are sustained across model sizes (Figure \ref{fig:main_transfer}) as well as across target domain annotation sizes (Figure \ref{fig:by_annot_size}). Furthermore, the gains from different source dimensions do not vary much, with the lowest average relative AUC gain of 6.74\% for SBIC ``Group'' and the highest of 13.46\% for HASOC ``Target''. For the negative transfer (\#MeToo ``Toxicity''), the impact with the smallest model (\mOne) is mixed, with transfer from SBIC ``Lewd'' offering a small mean gain of 4.67\%, while transfer from SBIC ``Group'' results in a small mean loss of -1.30\%. It is worth noting that these two annotation dimensions are the least correlated on SBIC dataset (r=0.10). This mixed impact turns into mean loss of -3.30\% with an intermediate size model (\mEight{}) and a minor gain of 0.94\% with the largest model (\mTwenty). The transfer impact for \mTwenty{} varies between -3.91\% loss for SBIC ``Intent'' to 4.74\% gain for HASOC ``HOF''. Comparing the two scenarios, we can also see that the initial baseline performance of the models is consistently higher for the negative transfer scenario (mean AUC of 58.9) than for the positive one (mean AUC of 55.6). 

\paragraph{Active Learning Effectiveness:} Looking at average relative gain across annotation sizes in Figure \ref{fig:sample_size_transfer}, we can see that without any annotated target samples (i.e., no active learning), the gains from transfer are small, but appear in both scenarios (4.6\% for ``Sexually Explicit'' and 2.7\% for ``Toxicity''). Annotation of just 100 target samples differentiates the scenarios leading to big average gain of 16.0\% for positive transfer and to a small average loss of -2.6\% for the negative transfer. Looking at absolute AUCs in Figure \ref{fig:by_annot_size} we can observe that mixing small number of target domain samples within transfer regime, can lead to large AUC gain from 56.6 to 63.5 (12.1\%). This is comparing transfer without any target annotations and with just 100 annotations respectively. We further observe that as annotation size increases, the relative gain from using external data decreases by 26.10\% from 100 to 2k annotated target samples (Figure \ref{fig:sample_size_transfer}). The largest drop (20.48\% decrease in relative AUC gain) takes place between 100 and 1k annotated examples, which is also a 10-fold increase in the size of annotated target data. Annotation of additional 1k examples, representing just a 2-fold increase, leads to a much smaller impact (7.07\% decrease in gain). We can also see that higher proportion of target-domain samples are used as shots as annotation size increases \ref{fig:shot_ratio}. Finally, we observe that active learning alone provides small gains from AUC of 54.3 for zero-shot to 56.8 for 2k target annotations (relative gain of 4.9\%) for ``Sexually Explicit'' and form 58.5 for zero-shot to 59.2 for 2k annotations (relative gain of 1.3\%) for ``Toxicity''.

The main takeaways from these results are that: 1) if the positive or negative transfer occurs, it is retained across model and target annotation sizes, 2) the higher initial baseline AUC for the models likely contributes to the negative transfer, 3) transfer effectiveness can increase with small target domain annotation size, but diminishes with an increasing number of annotations.

\paragraph{Correlations between datasets and labeling properties:}
We perform additional analysis to understand the nature of positive and negative transfers. First, we examine whether the sheer amount of external data from the source domain impacts transfer effectiveness. We find that the smaller HASOC dataset (6k) actually offers a higher mean gain of 7.54\% compared to a much larger SBIC (34k) offering a mean gain of 4.76\% in the same setup. It is worth noting that we add these to our support set in their entirety, but the TF-IDF shot selection still picks the most relevant examples from this pool. We find that the difference in label imbalance between the source and target datasets is not correlated with AUC gain from the transfer ($\rho$=$0.14$). We also find that correlation between source and target labeling dimensions estimated on the source dataset (i.e., SBIC or HASOC) is only weakly related to AUC gain ($\rho$=$-0.27$). We find, however, that the higher the initial performance of the PLM with a given annotation size (i.e., without source domain data) the lower the AUC gain from the transfer ($\rho$=$-0.66$). 
Finally, we estimate the distributional difference wrt. labels between the datasets following the approach from \cite{zhao2021active}. We train an SVM classifier to tell datasets apart under the aligned labels (i.e., positive class posts put into the same set) from source and target domain tasks (separability in Figure \ref{fig:datasets_distribution_difference}), but we find only a weak correlation to the AUC gain ($\rho$=$0.25$).

\section{Discussion and Future Work}
\label{sec:discussion}

\paragraph{Impact of model size:} We observe that as the model size increases, the gains in positive transfer tend to decrease only slightly (2.45\% gap between gain from \mOne{} and \mTwenty{}) and the overall effectiveness of transfer is largely retained (Figure \ref{fig:main_transfer}). In a negative transfer scenario, however, small gains can be inconsistent and turn into losses  (1.91\% gain in \mOne{}, -3.3\% loss for \mEight{}, and a 0.9\% gain for \mTwenty{}). It is well documented that as the model size increases its capabilities on standard NLP tasks tend to increase \cite{min2022rethinking}. While the better performance and less need for external data of larger models are not surprising, the difference in the performance for different tasks may suggest that larger models may not gain capabilities uniformly (i.e., a large model may become much better at detecting ``Toxicity'', but improves only slightly in detecting ``Sexually Explicit'' content). 

\paragraph{Impact of annotation size:} As reported in the results, as annotation size increases, the relative gain from transfer decreases (Figure \ref{fig:sample_size_transfer}). The decrease in gain is due to the support set increasingly containing a higher proportion of target examples. Hence, these target examples are more likely to be used as shots as can be seen in Figure \ref{fig:shot_ratio}. In effect, the performance will approach the baseline (where all the shots are from the target domain). In our shot selection, we are currently not controlling for the proportion of source and target domain documents being used (i.e., we only balance labels). An additional set of experiments could explore label and domain-balanced shots selection, which could mitigate this behavior.

\paragraph{Understanding positive \& negative transfers:} Our results suggest that negative transfer is more likely to happen if 1) the initial PLM baseline on that task is higher and 2) the source dataset supplies examples that provide little new information on top of the already used target data. The first reason is intuitive, as a higher baseline is harder to beat. The initial high baseline also reflects how well the PLMs internal knowledge already informs the target-domain task. The second finding is currently only anecdotal (correlations are weak) and much less intuitive. It should also be interpreted within the space of datasets used for our experiments. Taken at face value, it suggests that the more different the data, the higher the gain from the transfer, which is unlikely to be true. While our datasets and labels are different at the task level (i.e., ``Lewd'' content is likely slightly different than ``Sexually Explicit'' content), they also represent a similar broader domain of hate speech, toxicity, and stereotypical bias on social media. In that sense, they come from a similar domain and capture similar tasks (we report label similarities in Appendix \ref{sec:correlations_appendix} and distributional differences in Appendix \ref{sec:separability_similarity_appendix}). In this interpretation, we are likely observing the benefits of diversity and novelty of external data used for shots within the broader related domain, similar to the benefits of domain-adaptive pretraining \cite{gururangan2020don}. Future work should examine using source datasets coming from an entirely different broader domain (e.g., Enron email dataset \cite{EnronEma54:online}), which are unlikely to lead to positive transfer.

\paragraph{Limitations \& Practical application:} One limitation of our work is that the datasets we use rely on untrained crowd-sourced labeling which can be noisy and based on personal biases and perceptions \cite{binns2017like}. Perspective API labeling has known limitations of its own \cite{bender2021dangers}. Furthermore, PLMs can be biased and toxic themselves when prompted \cite{gehman2020realtoxicityprompts}, which also likely allows them to detect these dimensions based on their internal knowledge \cite{schick2021self}. Our proposed method can, unfortunately, be misused intentionally or unintentionally \cite{weidinger2021ethical}. 
We specifically see the dangers of using our approach for censorship \cite{ullmann2020quarantining}. Some future applications of our ATF method involve noisy pre-labeling of unlabeled datasets and selecting samples for future fine-tuning (e.g., via disagreement-based active learning \cite{hanneke2014theory}). With some limited initial human labeling of as few as $100$ random documents, if the baseline few-shot performance is poor, using prelabeled out-of-domain data can improve the AUC without expending more human annotation effort. We also plan to use this method to efficiently label custom dimensions of toxicity relevant to \#MeToo and other real-world data, which are currently not supported by tools such as Perspective API.

\section{Conclusion}

In this paper, we present ATF, a novel adaptation of few-shot instructions to facilitate transfer learning without fine-tuning on PLMs in a setting with limited labeling resources. We demonstrate that our method can lead to consistently high AUC gains across model and annotation sizes with a small amount of annotated data from the target dimension. We also observe positive and negative transfer scenarios and find that higher AUC of PLM without any pre-annotated source domain data is correlated with less gain in AUC from the transfer. Our results motivate future work in understanding when ATF is useful and how it can be improved, as well as practical applications including noisy pre-labeling and sample selection for fine-tuning. 

\begin{ack}
We would like to thank the Caltech SURF program for contributing to the funding of this project and especially the named donor Carolyn Ash. This material is based upon work supported by the National Science Foundation under Grant \# 2030859 to the Computing Research Association for the CIFellows Project. Anima Anandkumar is partially supported by Bren Named Chair Professorship at Caltech and is a paid employee of Nvidia. Sara Kangaslahti was a paid part-time intern at Nvidia during this project.
\end{ack}

\bibliographystyle{abbrv}


\appendix

\section{Appendix - Correlations between labels on \#MeToo, SBIC and HASOC datasets}
\label{sec:correlations_appendix}

\begin{figure}[H]
  \begin{minipage}{0.49\linewidth}
    \includegraphics[width=\linewidth]{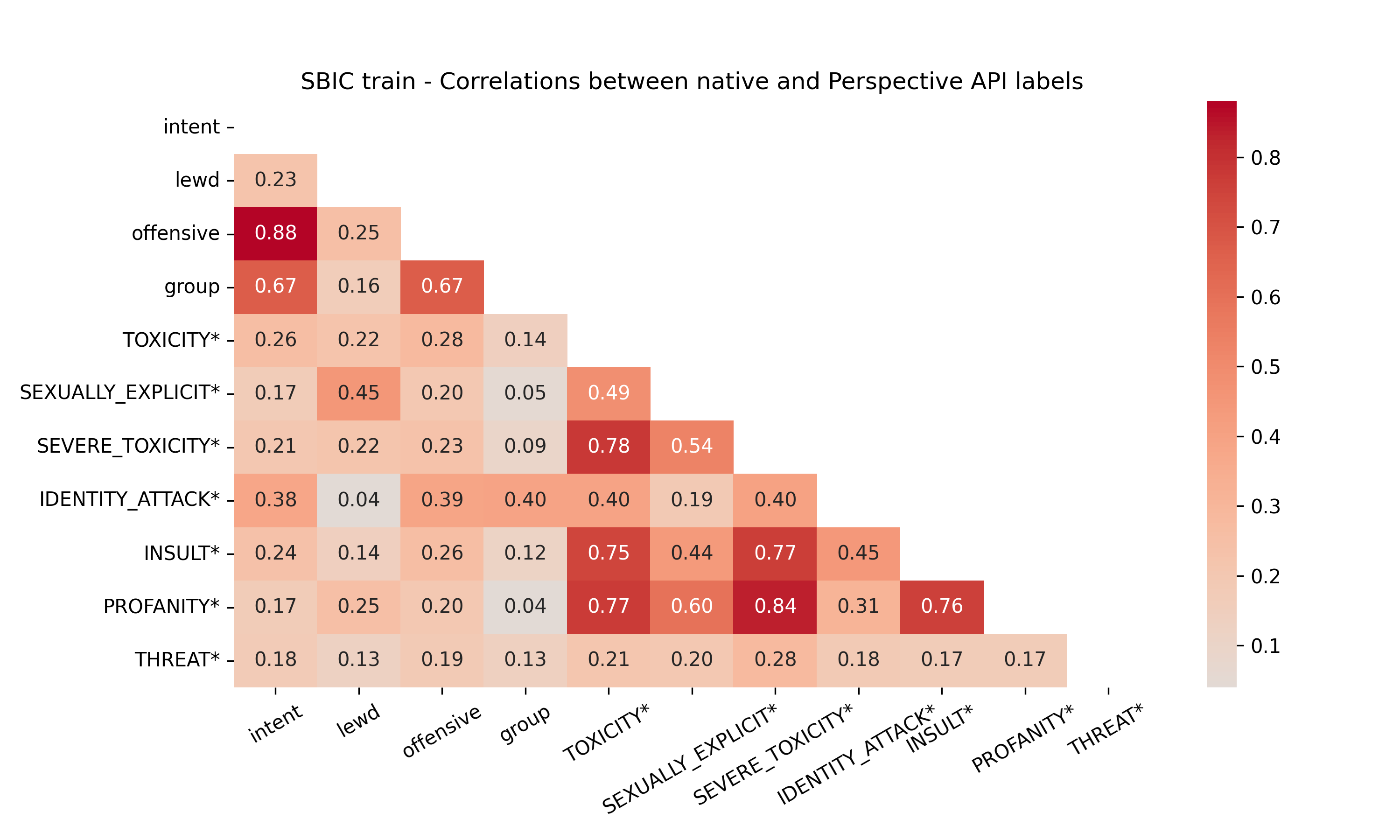}
  \end{minipage}
  \hfill
  \begin{minipage}{0.49\linewidth}
      \includegraphics[width=\linewidth]{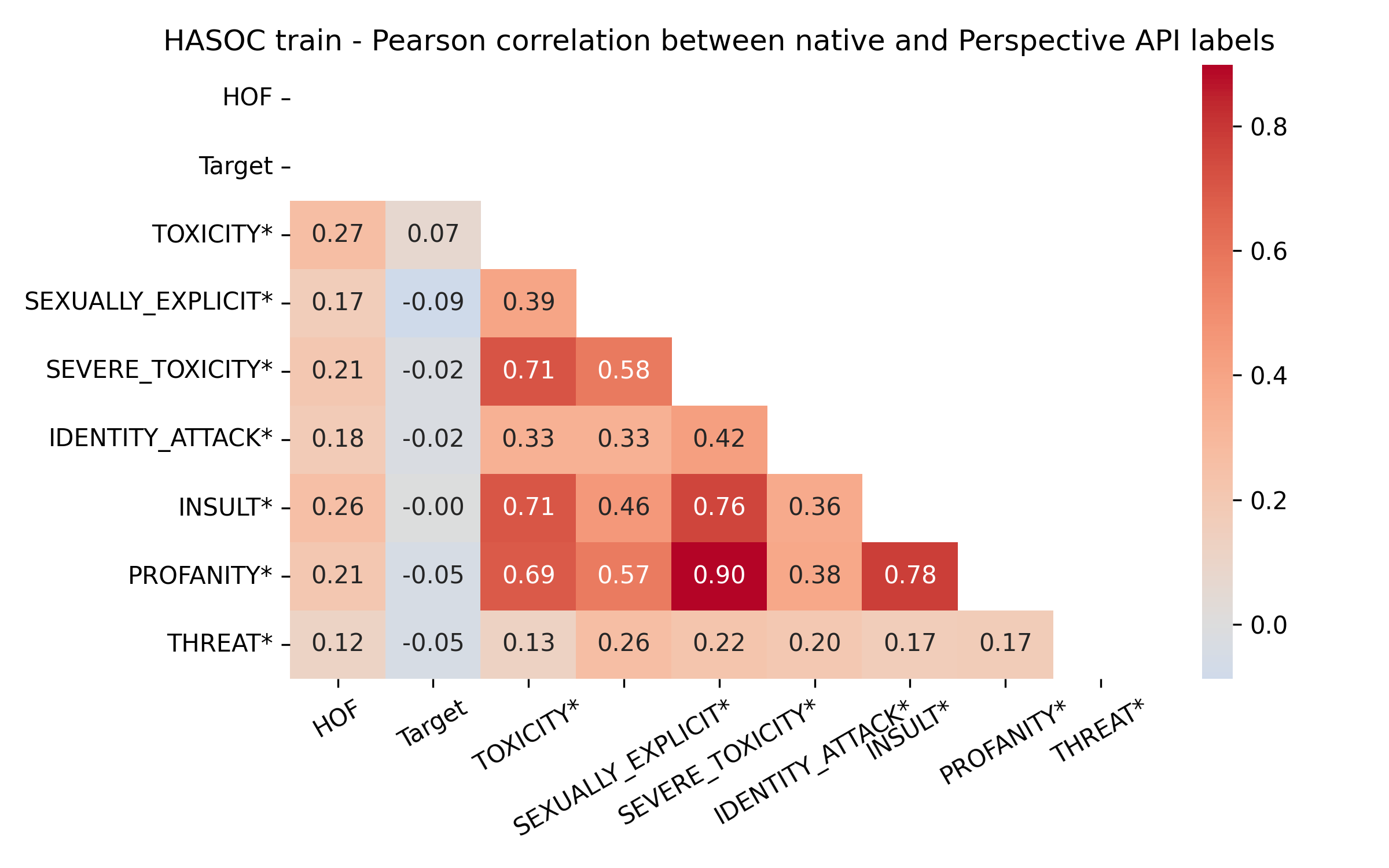}
  \end{minipage}
   \hfill
  \begin{minipage}{0.49\linewidth}
      \includegraphics[width=\linewidth]{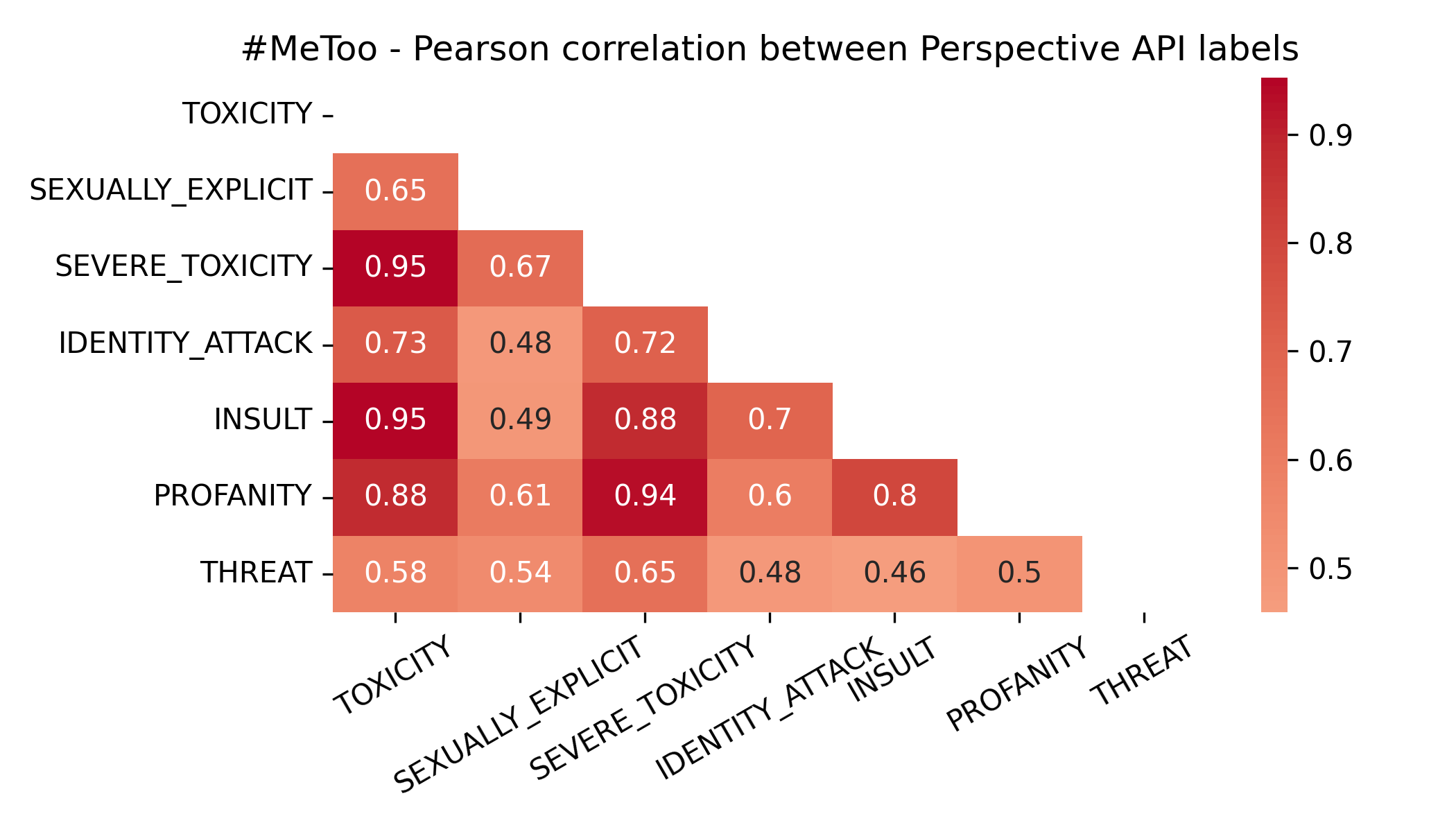}
  \end{minipage}
  \caption{Pearson Correlations between native labels and Perspective API labels on \#MeToo, SBIC, and HASOC datasets. Dimensions with ``*'' provided by Perspective API} 
  \label{fig:datasets_corrs}
\end{figure}

\section{Appendix - Separability \& Semantic similarity between the dataset-annotation pairs}
\label{sec:separability_similarity_appendix}

\begin{figure}[H]
  \begin{minipage}{1.0\linewidth}
    \includegraphics[width=\linewidth]{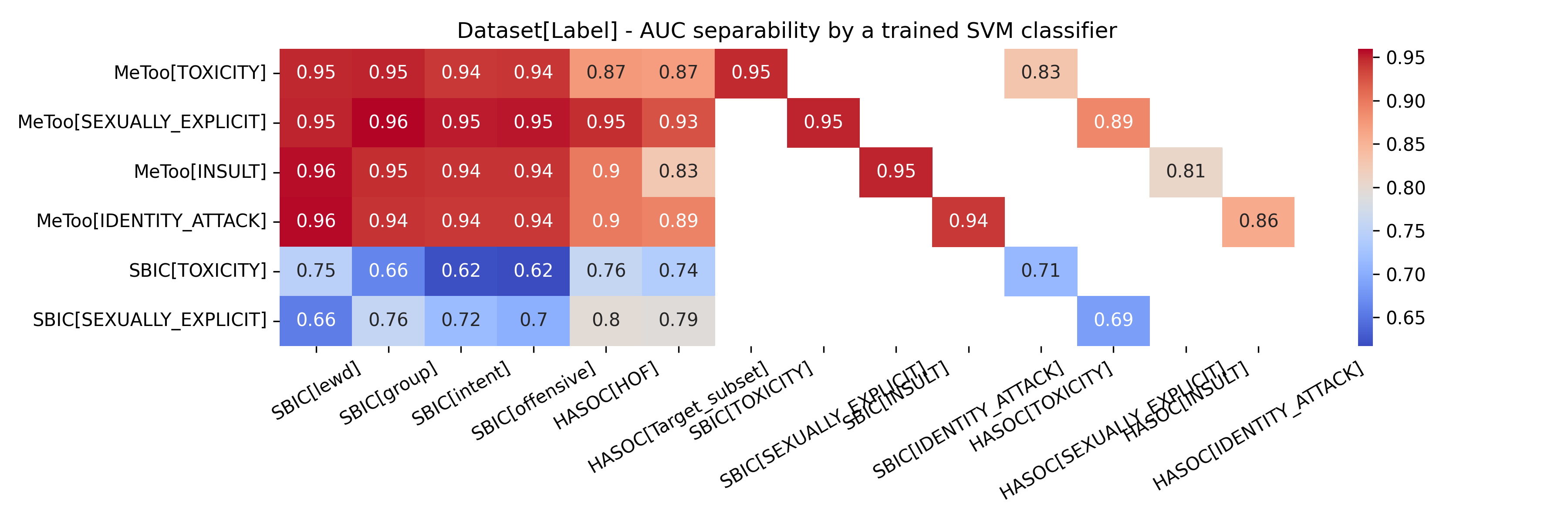}
  \end{minipage}
  \hfill
  \begin{minipage}{1.0\linewidth}
      \includegraphics[width=\linewidth]{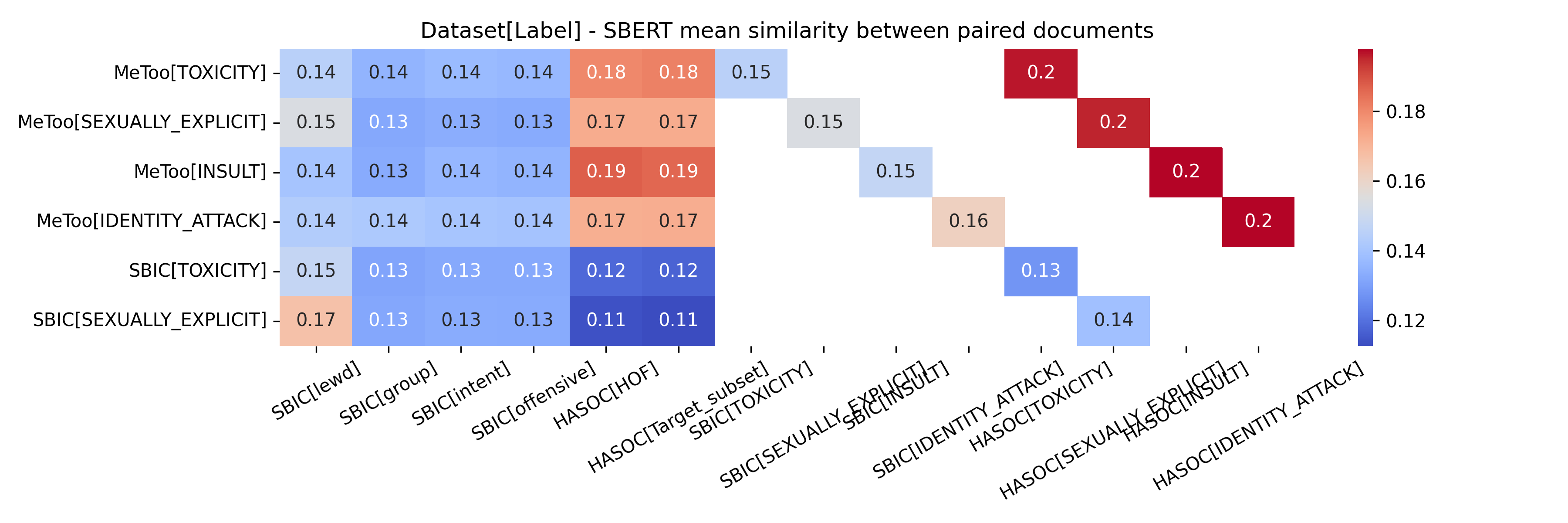}
  \end{minipage}
  \caption{Estimate of distributional difference - SVM separability (top) and SBERT mean similarity (bottom) between dataset-label pairs for \#MeToo, SBIC, and HASOC datasets. We estimate the distributional difference wrt. labels between the datasets following the approach from \cite{zhao2021active}. We train an SVM classifier to tell datasets apart under the aligned labels (i.e., positive class posts put into the same set) from the source and target domain tasks.} 
  \label{fig:datasets_distribution_difference}
\end{figure}

\section{Appendix - Labeling definitions used with few-shot instructions}
\label{sec:labeling_definition}

\begin{table*}[h]
\centering
\small{
\begin{tabular}{llll}
\textbf{Dataset} & \textbf{Task}                                                & \textbf{Definition}                                                                                                                                              & \textbf{Tokens} \\
\toprule
SBIC             & Offensive                                                    & Does this post contain offensive language?                                                                                                                       & Yes, No         \\
                 & Intent                                                       & Does this post contain intentional insults?                                                                                                                      & Yes, No         \\
                 & Lewd                                                         & Does this post contain sexual content?                                                                                                                           & Yes, No         \\
                 & Group                                                        & Does this post contain offense to a group?                                                                                                                       & Yes, No         \\
\midrule
HASOC            & HOF                                                          & Does this post contain any form of non-acceptable language such as & Yes, No \\
 & & hate speech, offensiveness, aggression, profanity? \\
                 & Target                                                       & Does this post contain an insult/threat to an individual, group, or others?                                                                                      & Yes, No         \\
\midrule
All datasets            & Toxicity                                                     & Does this post contain rude, disrespectful, or unreasonable language?                                                                                            & Yes, No         \\
                 & Sexually Explicit & Does this post contain sexually explicit language?                                                                                                               & Yes, No \\
\bottomrule
\end{tabular}
}
\vspace{1.0em}
\caption{Labeling definitions used for each task in the few-shot prompts. We use the definitions from prior few-shot work for the SBIC and HASOC tasks \cite{prabhumoye2021few}. For the Perspective API tasks we use the definitions of each dimension provided on the Perspective API website \cite{PerspectiveAPI:online}. We note that in all transfer learning experiments, we use the source-domain definition for the source-domain shots and the target-domain definition for the target-domain shots and query.}
\vspace{-1.0em}
\label{tab:labeling_definitions_table}
\end{table*}

\end{document}